\documentclass{article}
\usepackage{ijcai19}

\usepackage{times}
\usepackage{soul}
\usepackage{url}
\usepackage[hidelinks]{hyperref}
\usepackage[utf8]{inputenc}
\usepackage[small]{caption}
\usepackage{graphicx}
\usepackage{amsmath}
\usepackage{booktabs}
\usepackage{algorithm}
\usepackage{algorithmic}
\usepackage{helvet}
\usepackage{courier}
\usepackage{latexsym}
\usepackage{subfigure}
\usepackage{xcolor}
\usepackage{enumerate}
\usepackage{comment}
\usepackage{subfigure}
\usepackage{multirow}
\usepackage{bm}
\usepackage{wrapfig}

\title{Parts4Feature: Learning 3D Global Features from Generally Semantic Parts in Multiple Views}

\author{Zhizhong Han\textsuperscript{1,2}, Xinhai Liu\textsuperscript{1}, Yu-Shen Liu\textsuperscript{1}\thanks{Corresponding author: Yu-Shen Liu}, Matthias Zwicker\textsuperscript{2}\\
\textsuperscript{1}School of Software, Tsinghua University, Beijing, China \\
Beijing National Research Center for Information Science and Technology (BNRist)\\
\textsuperscript{2}Department of Computer Science, University of Maryland, College Park, USA\\
h312h@umd.edu,
lxh17@mails.tsinghua.edu.cn,
liuyushen@tsinghua.edu.cn,
zwicker@cs.umd.edu
}

\begin{document}

\maketitle

\begin{abstract}
Deep learning has achieved remarkable results in 3D shape analysis by learning global shape features from the pixel-level over multiple views. Previous methods, however, compute low-level features for entire views without considering part-level information. In contrast, we propose a deep neural network, called \textit{Parts4Feature}, to learn 3D global features from part-level information in multiple views. We introduce a novel definition of generally semantic parts, which Parts4Feature learns to detect in multiple views from different 3D shape segmentation benchmarks. A key idea of our architecture is that it transfers the ability to detect semantically meaningful parts in multiple views to learn 3D global features. Parts4Feature achieves this by combining a local part detection branch and a global feature learning branch with a shared region proposal module. The global feature learning branch aggregates the detected parts in terms of learned part patterns with a novel multi-attention mechanism, while the region proposal module enables locally and globally discriminative information to be promoted by each other. We demonstrate that Parts4Feature outperforms the state-of-the-art under three large-scale 3D shape benchmarks.
\end{abstract}

\section{Introduction}
Learning 3D global features from multiple views is an effective approach for 3D shape understanding. A widely adopted strategy is to leverage deep neural networks to aggregate features hierarchically extracted from pixel-level information in each view. However, current approaches can not employ part-level information. In this paper, we show for the first time how extracting part-level information over multiple views can be leveraged to learn 3D global features. We demonstrate that this approach further increases the discriminability of 3D global features and outperforms the state-of-the-art methods on large scale 3D shape benchmarks.


It is intuitive that learning to detect and localize semantic parts could help classify shapes more accurately. Previous studies on fine-grained image recognition also employ this intuition by combining local part detection and global feature learning together. To learn highly discriminative features to distinguish subordinate categories, these methods try to first detect important parts, such as heads, wings and tails of birds, and then collect these part features into a global feature.
However, these methods do not tackle the challenges that we are facing in the 3D domain.
First, these methods require ground truth parts with specified semantic labels, while 3D shape classification benchmarks do not provide such kind of labels.
Second, the part detection knowledge learned by these methods cannot be transferred for general purpose use, such as non-fine-grained image classification, since it is specified for particular shape classes. 
Third, these methods are not designed to aggregate part information from multiple images, corresponding to multiple views of a 3D shape in our scenario. Therefore, simultaneously learning
part detection and further aggregating part-level information from multiple views become a unique challenge in 3D global feature learning.

To address these issues, we propose \textit{Parts4Feature}, a deep neural network to learn 3D global features from semantic parts in multiple views. With a novel definition of generally semantic parts (GSPs), Parts4Feature learns
to detect GSPs in multiple views from different 3D shape segmentation benchmarks. 
Moreover, it learns a 3D global feature from shape classification data sets,
by transferring the learned knowledge of part detection, and leveraging the detected GSPs in multiple views. Specifically, Parts4Feature is mainly composed of a local part detection branch and a global feature learning branch.
Both branches share a region proposal module, which enables
locally and globally discriminative information to get promoted by each other.


The local part detection branch employs a novel neural network derived from Fast R-CNN~\cite{Girshick2015} to learn to detect and localize GSPs in multiple views. In addition, the global feature learning branch
incrementally aggregates the detected parts in terms of learned part patterns with multi-attention. We propose a novel multi-attention mechanism to further increase the discriminability of learned features by not only highlighting the distinctive parts and part patterns but also depressing the ambiguous ones.
Our novel view aggregation based on semantic parts prevents information loss caused by the widely used pooling, and it can understand each detected part in a more detailed manner. In summary, our contributions are as follows:

\begin{enumerate}[i)]
\item We propose Parts4Feature, a novel deep neural network to learn 3D global features from semantic parts in multiple views, by combining part detection and global feature learning together.
\item We show that the novel structure of Parts4Feature is capable of learning and transferring universal knowledge of part detection, 
which allows Parts4Feature to leverage discriminative information from another source (3D shape segmentation) for 3D global feature learning.
\item Our global feature learning branch introduces a novel view aggregation based on semantic parts, where the proposed multi-attention further improves the discriminability of learned features.
\end{enumerate}

\section{Related work}
\label{sec:relatedworks}
\noindent\textbf{Mesh-based deep learning models. }To directly learn 3D features from 3D meshes, different novel concepts, such as circle convolution~\cite{Zhizhong2016bijcai}, mesh convolution~\cite{Zhizhong2016ijcai} were proposed to perform in deep learning models. These methods aim to learn global or local features from the geometry and spatial information on meshes to understand 3D shapes.



\noindent\textbf{Voxel-based deep learning models. } Similar to images, voxels have regular structure to be learned by deep learning models, such as CRBM~\cite{Wu2015ijcai}, fully convolutional denoising autoencoders~\cite{Sharma16}, CNNs~\cite{su16mvcnn}, GAN~\cite{3dganWuijcai}. These methods usually employ 3D convolution to better capture the contextual information in local regions. Moreover, Tags2Parts~\cite{MuralikrishnanK18} discovered semantic regions that strongly correlate with user-prescribed tags by learning from voxels using a novel U-Net.



\noindent\textbf{Deep learning models for point clouds. }As a series of pioneering work, PointNet++~\cite{nipspoint17ijcai} inspired various supervised methods to understand point clouds. Through self-reconstruction, FoldingNet~\cite{YaoqingCVPR2018} and LatentGAN~\cite{PanosCVPR2018ICMLijcai} learned global features with different unsupervised strategies.

\noindent\textbf{View-based deep learning models. }Similar to the light field descriptor (LFD), GIFT~\cite{tmmbs2016ijcai} measured the difference between two 3D shapes using their corresponding view feature sets. Moreover, pooling panorama views~\cite{Bshi2015ijcai,Sfikas17ijcai} or rendered views~\cite{su15mvcnnijcai,Zhizhong2019seq} is more widely used to learn global features. Different improvements from camera trajectories~\cite{JohnsLD16}, view aggregation~\cite{chuwang2017,Zhizhong2018seqijcai}, pose estimation~\cite{AsakoCVPR2018} are also presented. However, these methods can not leverage part-level information.
In contrast, Parts4Feature learns and transfers universal knowledge of part detection to facilitate 3D global feature learning.



\section{Parts4Feature}

\begin{figure*}[tb]
  \centering
   \includegraphics[width=\linewidth]{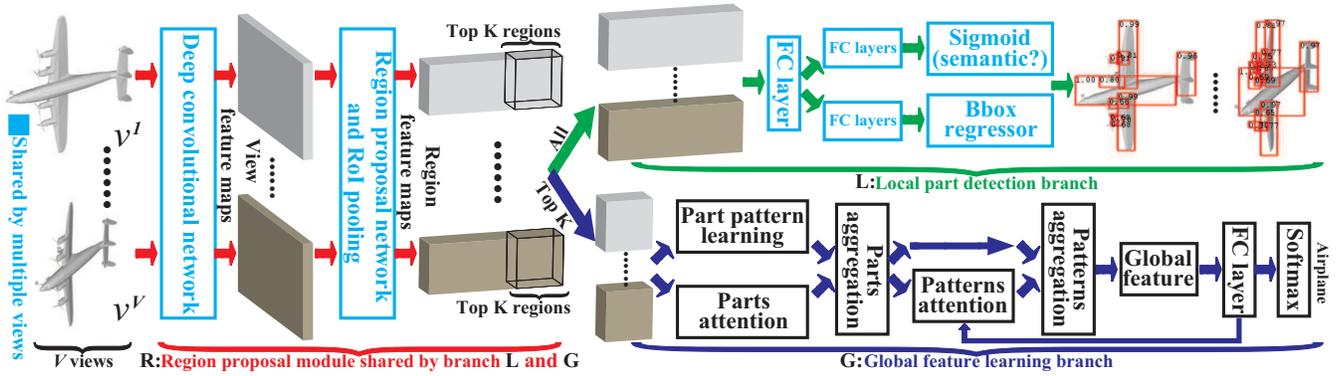}
  %
  %
\caption{\label{fig:framework} The demonstration of Parts4Feature framework.}
\end{figure*}

\noindent\textbf{Overview. }Parts4Feature consists of three main components as shown in Fig.~\ref{fig:framework}: a local part detection branch $\rm L$, a global feature learning branch $\rm G$, and a region proposal module $\rm R$, where $\rm R$ is shared by $\rm L$ and $\rm G$ and receives multiple views of a 3D shape as input. We train Parts4Feature simultaneously under a local part detection benchmark $\bm{\Phi}$ and a global feature learning benchmark $\bm{\Psi}$. The local part detection branch $\rm L$ learns to identify GSPs in multiple views under $\bm{\Phi}$, while $\rm G$ learns a global feature from the detected GSPs in multiple views under $\bm{\Psi}$.

For a 3D shape $m$ in either $\bm{\Phi}$ or $\bm{\Psi}$, we capture $V$ views $v^i$ around it, forming a view sequence $\bm{v}=\{v^i|i\in [1,V]\}$. First, the region proposal module $\rm R$ provides the features $\bm{f}_{j}^i$ of regions $r_{j}^i$ proposed in each view $v^i$, where $j\in[1,J]$. Then, by analyzing the region features $\bm{f}_{j}^i$ in $\bm{v}$, branch $\rm L$ learns to predict what and where GSPs are in multiple views. Finally, by aggregating the features $\bm{f}_{k}^i$ of the top $K$ region proposals $r^i_j$ in each $v^i$ in $\bm{v}$, the global feature learning branch $\rm G$ produces the global feature $\bm{f}$ of shape $m$. Our approach to aggregating region proposal features is based on $N$ semantic part patterns $\bm{\theta}_n$ with multi-attention for 3D shape classification, where
$\bm{\theta}_n$ are learned across all views in the global feature learning benchmark $\bm{\Psi}$.

\noindent\textbf{Generally semantic parts. }We define a GSP
as a local part in any semantic part category of any shape class, such as engines of airplanes or wheels of cars. Although our concept of GSPs simplifies all semantic part categories into a binary label by only determining whether a part is semantic or not, this allows us to exploit discriminative, part-level information from several different 3D shape segmentation benchmarks for global feature learning.

\begin{figure}[tb]
  \centering
   \includegraphics[width=\linewidth]{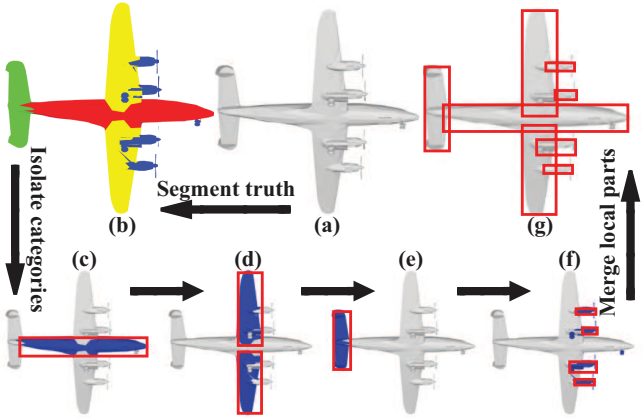}
  %
  %
\caption{\label{fig:parts} The illustration of ground truth GSPs construction.}
\end{figure}

We use three 3D shape segmentation benchmarks involved in~\cite{KalogerakisAMC17ijcai}, including ShapeNetCore,
Labeled-PSB, and COSEG to construct the local part detection benchmark $\bm{\Phi}$ and provide ground truth GSPs. We also split the 3D shapes in each segmentation benchmark into training and test sets according to~\cite{KalogerakisAMC17ijcai}. Fig.~\ref{fig:parts} shows the construction of ground truth GSPs. For each view $v^i$ of a 3D shape $m$ shown in Fig.~\ref{fig:parts}(a), we obtain its ground truth segmentation visualized in Fig.~\ref{fig:parts}(b) from the shape segmentation benchmark. Then, we can isolate each part category to precisely locate GSPs, as shown from Fig.~\ref{fig:parts}(c) to Fig.~\ref{fig:parts}(f). We emphasize each isolated part category in blue, where we locate the corresponding GSPs by computing the bounding box (red) of the colored regions. Finally, we show all GSPs in view $v^i$ in Fig.~\ref{fig:parts}(g). We collect all GSPs of shape $m$ by repeating these procedures in all its $V$ views. Note that we omit small GSPs (for example the landing gear in Fig.~\ref{fig:parts}(f)) whose bounding boxes are smaller than 0.45 of the max bounding box in the same part category.

\noindent\textbf{Region proposal module $\rm R$. }$\rm R$ provides region proposals $r_j^i$ in all views $v^i$ and their features $\bm{f}^i_j$, which are then forwarded to the local part detection and global feature learning branches. Shared by all views $v^i$ in $\bm{v}$, $\rm R$ is composed of a Deep Convolutional Network (DCN), and a Region Proposal Network (RPN) with Region of Interest (RoI) pooling~\cite{Girshick2015}.

DCN is modified from a VGG$\_$CNN$\_$M$\_$1024 network~\cite{ChatfieldSVZ14}, and it produces  feature $\bm{f}^i$ for each view $v^i$ as 512 feature maps of size $12\times 12$. Based on $\bm{f}^i$, RPN then proposes regions $r^i_j$ in a sliding-window manner. At each sliding-window location centered at each pixel of $\bm{f}^i$, a region $r^i_j$ is proposed by determining its location $t_{\rm R}$ and predicting GSP probabilities $p_{\rm R}$ with an anchor. The location $t_{\rm R}$ is a four dimensional vector representing center, width and height of the bounding box. We use 6 scales and 3 aspect ratios to yield $6\times 3=18$ anchors, which ensures a wide range of sizes to accommodate region proposals for GSPs that may be partially occluded in some views. The 6 scales relative to the size of the views are $[1,2,4,8,16,32]$, and the 3 aspect ratios are $1:1$, $1:2$, and $2:1$. Altogether, this leads to $J=2592 = 12\times 12\times 18$ regions $r^i_j$ in each view $v^i$.

To train RPN to predict GSP probabilities $p_{\rm R}$, we assign a binary label to each region $r^i_j$ indicating whether $r^i_j$ is a GSP. We assign a positive label if the Intersection-over-Union (IoU) overlap between $r^i_j$ and any ground-truth GSP in $v^i$ is higher than a threshold $S_{\rm R}$, and we use a negative label otherwise.
In each view $v^i$ we apply RoI pooling over regions given by $\{t_{\rm R}\}$ on feature maps $\bm{f}^i$. Hence, the features $\bm{f}^i_j$ of all $J$ region proposals $r^i_j$ are $512\times 7\times 7$ dimensional vectors, which we forward to the local part detection branch $\rm L$. In addition, we provide the features $\bm{f}^i_k$ of the top $K$ regions $r^i_j$ according to their GSP probability $p_{\rm R}$ to the global feature learning branch $\rm G$.

\noindent\textbf{Local part detection branch $\rm L$. }The purpose of this branch is to detect GSPs from the $J$ region proposals $r^i_j$ in each view $v^i$. We employ $\rm L$ as an enhancer of RPN, where $\rm L$ aims to learn what and where GSPs are in $v^i$ without anchors in a more precise manner. The intuition is that this in turn pushes RPN to propose more GSP-like regions, which we provide to the global feature learning branch $\rm G$.

We feed the region features $\bm{f}^i_j$ of $r^i_j$ into a sequence of fully connected layers followed by two output layers. The first one estimates the GSP probability $p_{\rm L}$ that $r^i_j$ is a GSP using a sigmoid function as an indicator. The second one predicts the corresponding part location $t_{\rm L}$ using a bounding box regressor, where $t_{\rm L}$ represents the same bounding box parameters as $t_{\rm R}$ in RPN.
Similar to the threshold $S_{\rm R}$ in $\rm R$, $\rm L$ employs another threshold $S_{\rm L}$ to assign positive and negative labels for training. Denoting the ground truth probabilities and locations of positive and negative samples in RPN $\rm R$ as $p'$ and $t'$, and similarly for $\rm L$ as $p''$ and $t''$, the objective function of Parts4Feature for GSP detection is formed by the loss in $\rm R$ and $\rm L$, which is defined for each region proposal as follows,

\begin{equation}
\begin{aligned}
\label{eq:objective}
& O_{\rm R+\rm L}(p_{\rm R},p_{\rm L},p',p'',t_{\rm R},t_{\rm L},t',t'')= \\
& O_{p}(p_{\rm R},p')+\lambda O_{t}(t_{\rm R},t')+O_{p}(p_{\rm L},p'')+\lambda O_{t}(t_{\rm L},t''),\\
\end{aligned}
\end{equation}

\noindent where $O_{p}$ measures the accuracy in terms of GSP probability by the cross-entropy function of positive labels, while $O_{t}$ measures the accuracy in terms of location by the robust $L_1$ function as in~\cite{Ren2015ijcai}. The parameter $\lambda$ balances $O_{p}$ and $O_{t}$ in both $\rm R$ and $\rm L$. It works well in our experiments with a value of 1. In summary, Parts4Feature has the powerful ability to detect GSPs by simultaneously leveraging the view-level features $\bm{f}^i$ in $\rm R$ and the part-level features $\bm{f}^i_j$ in $\rm L$, which addresses the difficulty of GSP detection from multiple views caused by rotation and occlusion effects.

\noindent\textbf{Global feature learning branch $\rm G$. }This branch learns to map the features $\bm{f}^i_k$ of the top $K$ region proposals $r^i_k$ in each view $v^i$ in $\bm{v}$ to the 3D global feature $\bm{f}$. 
To avoid information loss caused by widely used pooling for aggregation, $\rm G$ incrementally aggregates all $V\times K$ region features $\bm{f}^i_k$ in terms of semantic part patterns $\bm{\theta}_n$ with multi-attention, where we learn the patterns $\bm{\theta}_n$  across all training data in the global feature learning benchmark $\bm{\Psi}$. The motivation for learning part patterns to aggregate regions is that the appearance of GSPs is so various that it would limit the discriminability of global features $\bm{f}$. Our multi-attention mechanism includes attention weights for view aggregation on the part-level and the part-pattern-level, denoted by $\bm{\alpha}$ and $\bm{\beta}$, respectively. Here, $\bm{\alpha}$ models how each of the $N$ patterns $\bm{\theta}_n$ weights each of the $V\times K$ regions $r^i_k$, while $\bm{\beta}$ measures how the final, global feature $\bm{f}$ weights each of the $N$ patterns $\bm{\theta}_n$.

Specifically, we employ a single-layer perceptron to learn $\bm{\theta}_n$, where $\bm{\theta}_n$ has the same dimension as $\bm{f}^i_k$. $\bm{\alpha}$ is a $(V\times K)\times N$ matrix, where each entry $\bm{\alpha}((i,k),n)$ is the attention paid to each of the $(V\times K)$ regions $r^i_k$ by the $n$-th pattern $\bm{\theta}_n$. $\bm{\alpha}((i,k),n)$ is measured by a softmax function as $exp(\bm{w}_n^T\bm{f}_k^i+b_n)/\sum_{n'\in [1,N]} exp(\bm{w}_{n'}^T\bm{f}_{k}^{i}+b_{n'})$. With $\bm{\alpha}$, we first aggregate all $(V\times K)$ region features $\bm{f}^i_k$ into a pattern specific aggregation $\bm{\varphi}_n$ in terms of each pattern $\bm{\theta}_n$ by computing $\sum_{i\in [1,V],k\in [1,K]} \bm{\alpha}((i,k),n) (\bm{\theta}_n-\bm{f}^i_k)$. Then, we further aggregate all $N$ pattern specific aggregations $\bm{\varphi}_n$ into the final, global feature $\bm{f}$ of 3D shape $m$. This is performed by linear weighting with the $N$ dimensional vector $\bm{\beta}$, such that $\bm{f}=\sum_{n\in[1,N]}\bm{\beta}(n)\bm{\varphi}_n$. For clarity of exposition, we explain the details of how we obtain $\bm{\beta}$ further below.

\begin{table*}
  \caption{The effects of $S_{\rm R}$ and $S_{\rm L}$ on the performance of Parts4Feature under ModelNet40.}
  \label{table:threshold}
  \centering
  \begin{tabular}{cccccccc}
    \hline
     Metrics & $(0.7,0.5)$ & $(0.7,0.6)$ & $(0.7,0.7)$ & $(0.7,0.8)$ & $(0.5,0.5)$ & $(0.8,0.5)$ & $(0.6,0.6)$ \\
    \hline
     mAP & \textbf{77.28} & 69.35 & 66.12 & 56.97 & 75.39 & 72.32 & 69.51
    \\
     Acc & \textbf{93.40} & 93.15 & 92.38 & 92.50 & 92.67 & 92.71 & 92.95
    \\
    \hline
  \end{tabular}
\end{table*}

Finally, we use $\bm{f}$ to classify $m$ into one of $C$ shape classes by a softmax function after a fully connected layer, where the softmax function outputs the classification probabilities $\bm{p}$, such that each probability $\bm{p}(c)$ is defined as $exp(\bm{u}_c^T\bm{f}+a_c)/\sum_{c'\in[1,C]}exp(\bm{u}_{c'}^T\bm{f}+a_{c'})$. The objective function of $\rm G$ is the cross entropy between $\bm{p}$ and the ground truth probability $\bm{p}'$,

\begin{equation}
\label{eq:objective1}
O_{\rm G}(\bm{p},\bm{p}')=-\sum\nolimits_{c\in[1,C]}\bm{p}'(c)log\bm{p}(c).
\end{equation}

The intuition behind modelling part-pattern-level attention is to enable Parts4Feature to weight the pattern specific aggregations $\bm{\varphi}_n$ according to the 3D shape characteristics that it has learned. This leads Parts4Feature to differentiate shapes in detail. To implement this, $\bm{\beta}$ is designed to capture the similarities between each of the $N$ pattern specific aggregations $\bm{\varphi}_n$ and the $C$ shape classes. To represent the characteristics of $C$ shape classes, we propose to employ the weights $\bm{u}_c$ in the fully connected layer before the last softmax function, as illustrated in Fig.~\ref{fig:framework}. We first project $\bm{\varphi}_n$ and $\bm{u}_c$ into a common space using matrices $\bm{W}_1$ and $\bm{W}_2$. Then we compute normalized similarities using a linear mapping with $\bm{w}$ and $\bm{g}$ as follows, $\bm{\beta}=softmax((\bm{W}_1[\bm{\varphi}_n^T]_N+\bm{W}_2[\bm{u}_c^T]_C)\bm{w}+\bm{g})$, where learnable parameters $\bm{W}_1$ and $\bm{W}_2$ are $N\times N$ and $N\times C$ dimensional matrices, $\bm{w}$ and $\bm{g}$ are $d$ and $N$ dimensional vectors, $[\bullet]_\circ$ means stacking all $\circ$ vectors $\bullet$ into a matrix row by row.

\begin{figure}[tb]
  \centering
   \includegraphics[width=\linewidth]{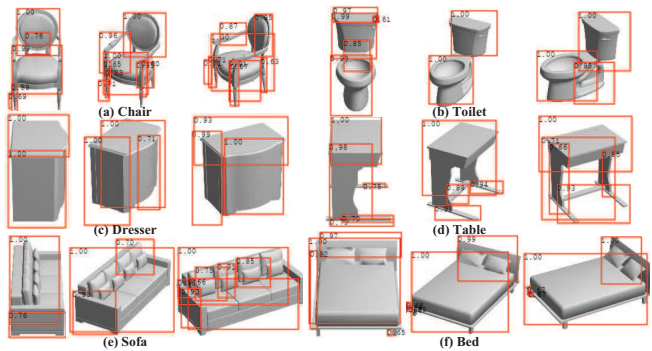}
  %
  %
\caption{\label{fig:detectionvis} The detected GSPs with $p_{\rm L}>0.6$ in red boxes.}
\end{figure}

\noindent\textbf{Training. }We train $\rm R$ and $\rm L$ together under a local part detection benchmark $\bm{\Phi}$, and $\rm G$ under a global feature learning benchmark $\bm{\Psi}$. The Parts4Feature objective is to simultaneously minimize Eq.~\ref{eq:objective} and Eq.~\ref{eq:objective1}, which leads to the loss

\begin{equation}
\begin{aligned}
\label{eq:objective2}
& O=\frac{1}{\|\bm{\Phi}\|}\sum\nolimits_{(p,t)\in\bm{\Phi}}O_{\rm R+\rm L}(p_{\rm R},p_{\rm L},p',p'',t_{\rm R},t_{\rm L},t',t'') \\
& +\frac{\eta}{\|\bm{\Psi}\|}\sum\nolimits_{\bm{p}\in\bm{\Psi}}O_{\rm G}(\bm{p},\bm{p}'),\\
\end{aligned}
\end{equation}

\noindent where the number of samples $\| \cdot \|$ is a normalization factor and $\eta$ is a balance parameter. Since $\rm R$ and $\rm L$ are based on the object detection architecture of Fast R-CNN~\cite{Girshick2015}, we adopt the approximate approach in~\cite{Ren2015ijcai} to jointly train $\rm R$ and $\rm L$ fast. In addition, we simultaneously update $\bm{u}_c$ in the softmax classifier in $\rm G$ by $\partial O_{\rm G}/\partial \bm{u}_c$ and $\partial\bm{\beta}/\partial\bm{u}_c$. This enables $\bm{u}_c$ to be learned more flexibly for optimization convergence, which is a connection across $\rm G$. For the $\bm{\Phi}=\bm{\Psi}$ case, parameters in $\rm R$, $\rm L$ and $\rm G$ can be simultaneously updated, otherwise, they are updated alternatively. For example, parameters in $\rm R$ and $\rm L$ are first updated under $\bm{\Phi}$, then, parameters in $\rm R$ (except RPN) and $\rm G$ are updated under $\bm{\Psi}$, and this process is iterated until convergence. 
In our following experiments we use $\eta=1$ .

\section{Experiments and analysis}

\begin{table*}[t]
  \caption{The effects of $K$, $N$ and $d$ on the performance of Parts4Feature under ModelNet10.}
  \label{table:parameters}
  \centering
  \begin{tabular}{cccccccccc}
    \hline
     Metric &$S_{\rm L}=0.5$&$0.7$&$0.8$& $K=10$&$30$&$N=128$&$512$&$V=3$&$6$\\
    \hline
     Acc &95.26 & \textbf{96.15} & 94.38& 94.38 & 94.93 & 94.49 & 95.04 & 94.27 & 94.93
    \\
    \hline
  \end{tabular}
\end{table*}


\noindent\textbf{Parameters. }We investigate how some important parameters affect Parts4Feature in shape classification under ModelNet~\cite{Wu2015ijcai}.

We first explore the IoU thresholds $S_{\rm R}$ in $\rm R$ and $S_{\rm L}$ in $\rm L$ that are used to establish positive GSP samples using ModelNet40~\cite{Wu2015ijcai} as $\bm{\Psi}$, as shown in Table~\ref{table:threshold}, where we initially use $V=12$ views, $K=20$ regions, and $N=256$ patterns. With $S_{\rm R}=0.7$ and increasing $S_{\rm L}$ from 0.5 to 0.8, the mean Average Presicion (mAP) under the test set of $\bm{\Phi}$ decreases, and accordingly, the average instance accuracy under the test set of $\bm{\Psi}$ decreases, compared to the highest classification accuracy $93.40\%$. With $S_{\rm L}=0.5$, we also decrease $S_{\rm R}$ to 0.5 and increase it to 0.8 respectively. The mAP only slightly drops from 77.28 to 75.39 and 72.32, although the corresponding accuracy decreases too. However, the mAP and the accuracy are not strictly positive correlated, as shown by ``(0.6,0.6)'', which has lower mAP but higher accuracy than ``(0.8,0.5)'' and ``(0.5,0.5)''. This comparison also implies that $S_{\rm L}$ affects part detection more than $S_{\rm R}$.



Next, we apply the parameters setting  ``(0.7,0.5)'' under ModelNet10~\cite{Wu2015ijcai}, as shown by the first accuracy of $95.26\%$ in Table~\ref{table:parameters}. Increasing $S_{\rm L}$ to 0.7 leads to an even better result of $96.15\%$. We also find the slight effect of $K$, $N$, and $V$ on the performance.

\begin{table}
  \caption{The view aggregation and attention comparison.}
  \label{table:aggregation}
  \centering
  \begin{tabular}{cc|cc}
    \hline
    \multicolumn{2}{c|}{Pooling} & \multicolumn{2}{|c}{Attention}     \\              \hline
    Methods & Acc & Methods & Acc \\
    \hline
    MaxPool & 90.97 & NoAtt & 92.84 \\
    MeanPool & 92.18 & PtAtt & 93.17 \\
    NetVLAD & 93.50 & PnAtt & 93.72 \\
    No $\rm L$ & 93.61 & MultiAtt & \textbf{96.15} \\
    \hline
  \end{tabular}
\end{table}

\begin{figure}[tb]
  \centering
   \includegraphics[width=\linewidth]{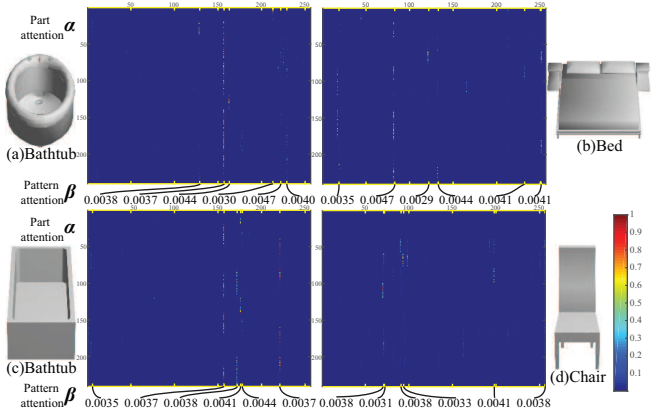}
  %
  %
\caption{\label{fig:attention} Multi-attention is visualized under the test set of ModelNet10. $\bm{\alpha}$ and $\bm{\beta}$ are shown as matrices and numbers.}
\end{figure}

We visualize part detection and multi-attention involved in our best result under ModelNet10 in Fig.~\ref{fig:detectionvis} and Fig.~\ref{fig:attention}, respectively. Although there are no ground truth GSPs under ModelNet10, Parts4Feature still successfully transfers the part detection knowledge learned from $\bm{\Phi}$ to detect GSPs in multiple views. Moreover, $\bm{\beta}$ is learned to focus on the patterns with high part attentions in $\bm{\alpha}$, where the top-6 patterns with high part attentions in $\bm{\alpha}$ are shown below for clarity.


\noindent\textbf{Ablation study. }Finally, in Table~\ref{table:aggregation} we highlight our semantic part based view aggregation and multi-attention method in branch $\rm G$ under ModelNet10. We replace our view aggregation with max pooling, mean pooling, and NetVLAD, where we aggregate $V\times K$ region features $\bm{f}^i_k$ for classification. Although these results are good, our novel aggregation with multi-attention can further improve the results. For evaluating multi-attention, we keep $\rm G$ unchanged and set all entries in $\bm{\alpha}$ and $\bm{\beta}$ to 1 (``NoAtt''). This leads to significantly worse performance compared to our ``MultiAtt''. Next, we employ $\bm{\alpha}$ and $\bm{\beta}$ separately. We find that both of part attention and part pattern attention improve ``NoAtt'', but $\bm{\alpha}$ (``PtAtt'') contributes less than $\bm{\beta}$ (``PnAtt''). Moreover, we highlight the effect of branch $\rm L$ as an enhancer of module $\rm R$ by removing $\rm L$ (``No $\rm L$'') from Parts4Feature, which is also justified by the degenerated results.

\begin{figure}[tb]
  \centering
   \includegraphics[width=\linewidth]{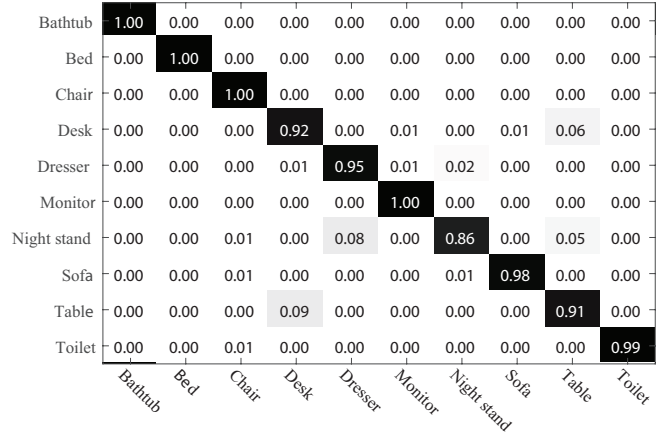}
  %
  %
\caption{\label{fig:MNConfusion} The classification confusion matrix under ModelNet10.}
\end{figure}

\noindent\textbf{Classification. }Table~\ref{table:comparison} compares Parts4Feature with the state-of-the-art in 3D shape classification under ModelNet. The comparison are conducted under the same condition\footnote{\noindent We use the same modality of views from the same camera system for the comparison, where the results of RotationNet are from Fig.4 (a) and (b) in https://arxiv.org/pdf/1603.06208.pdf. Moreover, the benchmarks are with the standard training and test split.}.

\begin{table}
  \caption{The classification comparison ModelNet.}
  \label{table:comparison}
  \centering
  \begin{tabular}{llll}
    \hline
    Methods & Raw  & MN40 & MN10 \\
    \hline
    MVCN\cite{su15mvcnnijcai} & View & 90.10 & - \\
    MVVC\cite{su16mvcnn} & Voxel & 91.40 & -\\
    3DDt\cite{JianwenCVPR2018ijcai} & Voxel & - & 92.40 \\
    PaiV\cite{JohnsLD16} & View & 90.70 & 92.80\\
    Sphe\cite{huang2017spherical} & View & 93.31 & - \\
    GIFT\cite{tmmbs2016ijcai} & View & 89.50 & 91.50 \\
    RAMA\cite{Sfikas17ijcai} & View & 90.70 & 91.12 \\
    VRN\cite{Brocknips2016}& Voxel & 91.33 &93.80 \\
    RNet\cite{AsakoCVPR2018}& View & 90.65 & 93.84 \\
    PNetP\cite{nipspoint17ijcai} & Point & 91.90 & -   \\
    DSet\cite{chuwang2017}& View & 92.20 & -\\
    VGAN\cite{3dganWuijcai}& Voxel & 83.30& 91.00 \\
    LAN\cite{PanosCVPR2018ICMLijcai} & Point & 85.70 & 95.30 \\
    FNet\cite{YaoqingCVPR2018} & Point& 88.40 & 94.40\\
    SVSL\cite{Zhizhong2018seqijcai}& View  & 93.31 & 94.82 \\
    VIPG\cite{Zhizhong2018VIPijcai}& View & 91.98 & 94.05 \\
    \hline
    Our & View & \textbf{93.40} & \textbf{96.15} \\
    \hline
  \end{tabular}
\end{table}

Under both benchmarks, Parts4Feature outperforms all its competitors at the same condition, where ``Our'' are obtained with the parameters of our best accuracy under ModelNet40 in Table~\ref{table:threshold} and the ones under ModelNet10 in Table~\ref{table:parameters}. This comparison shows that Parts4Feature effectively employs part-level information to significantly improve the discriminability of learned features. Parts4Feature is also outperforming under ShapeNet55 with the same parameters of our best results under ModelNet10, as shown by the comparison in the last three rows in Table~\ref{table:t10}.

\begin{figure}[tb]
  \centering
   \includegraphics[width=\linewidth]{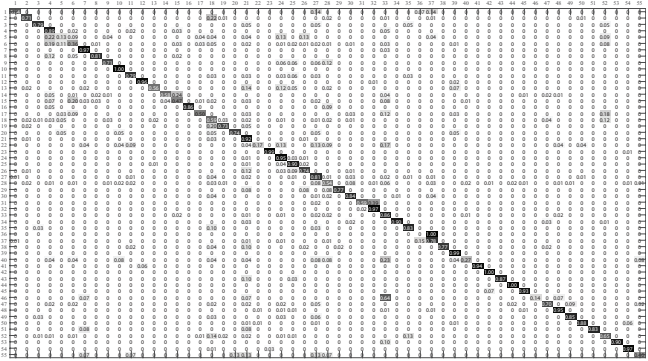}
  %
  %
\caption{\label{fig:SNConfusion} The classification confusion matrix under ShapeNet55.}
\end{figure}

To better demonstrate our classification results, we visualize the confusion matrix of our classification result under ModelNet10 and ShapeNet55 in Fig.~\ref{fig:MNConfusion} and Fig.~\ref{fig:SNConfusion}, respectively. In each confusion matrix, an element in the diagonal line means the classification accuracy in a class, while other elements in the same row means the misclassification accuracy. The large diagonal elements shows that Parts4Feature is good at classifying large-scale 3D shapes.

We also conduct experiments with reduced number of segmented shapes for training under ModelNet10. As shown in Table~\ref{table:reduce}, trained by randomly sampled $\{0\%,1\%,5\%,10\%,25\%,50\%\}$ of 6,386 shapes, our results increase accordingly. The good results with $0\%$ segmented shapes show that we not only learn from pixel-level information in 3D classification benchmarks, similar to existing methods, but also improve performance further by absorbing part-level information from 3D segmentation benchmark.

\begin{table}
  \caption{The effect of less segmented shapes for training.}
  \label{table:reduce}
  \centering
  \begin{tabular}{cccccccc}
    \hline
     Acc & $0\%$ & $1\%$ & $5\%$& $10\%$ & $25\%$ & $50\%$ & $100\%$\\
    \hline
     Instance  & 93.0  & 93.5 & 93.8 & 93.8 & 94.1 & 94.3 & \textbf{96.15}
    \\
     Class & 92.7  & 93.1 & 93.4 & 93.6 & 94.0 & 94.1 & \textbf{96.14}
    \\
    \hline
  \end{tabular}
\end{table}


\begin{figure}[tb]
  \centering
   \includegraphics[width=\linewidth]{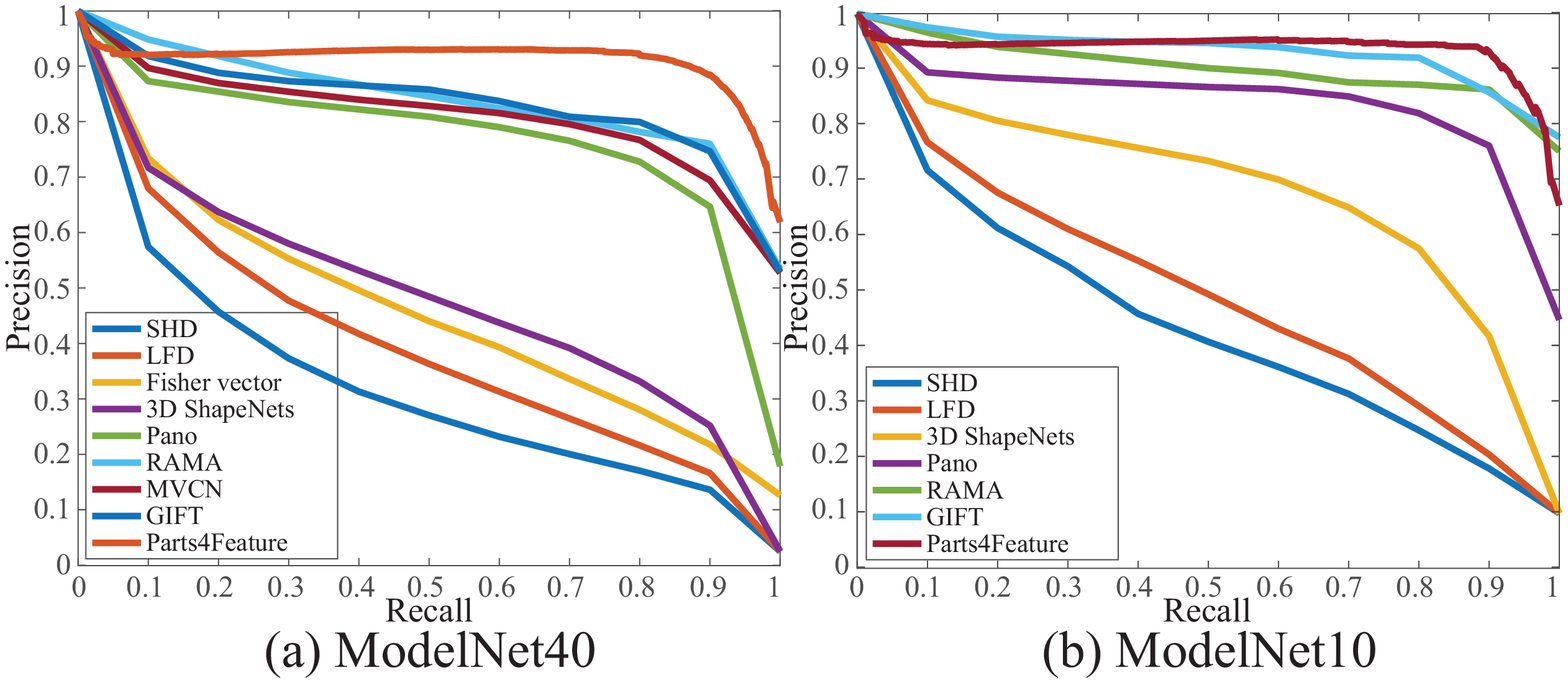}
  %
  %
\caption{\label{fig:retrieval} The PR curve comparison under ModelNet.}
\end{figure}




\noindent\textbf{Retrieval. }We further evaluate Parts4Feature in shape retrieval under ModelNet and ShapeNetCore55 by comparing with the state-of-the-art methods in Table~\ref{table:retrieval} and Table~\ref{table:t10}. These experiments are conducted under the test set, where each 3D shape is used as a query to retrieve from the rest of the shapes, and the retrieval performance is evaluated by mAP. The compared results include LFD, SHD, Fisher vector, 3D ShapeNets~\cite{Wu2015ijcai}, Pano~\cite{Bshi2015ijcai}, MVCN~\cite{su15mvcnnijcai}, GIFT~\cite{tmmbs2016ijcai}, RAMA~\cite{Sfikas17ijcai} and Trip~\cite{Xinweicvpr18} under ModelNet.

As shown in Table~\ref{table:retrieval}, Table~\ref{table:t10}, our results outperform all the compared results in each benchmark. Besides Taco~\cite{s2018spherical} in Table~\ref{table:t10}, the compared micro-averaged results in Table~\ref{table:t10} are from SHREC2017 shape retrieval contest~\cite{3dor20171050ijcai} with the same names. In addition, the available PR curves under ModelNet40 and ModelNet10 are also compared in Fig.~\ref{fig:retrieval}, which also demonstrates our outperforming results in shape retrieval.



\begin{table}
  \caption{The retrieval (mAP) comparison under ModelNet.}
  \label{table:retrieval}
  \centering
  \begin{tabular}{ccccccc}
    \hline
     Data & Pano & MVCN & GIFT& RAMA & Trip & Ours\\
    \hline
     MN40  & 76.8  & 79.5 & 81.9 & 83.5 & 88.0 & \textbf{91.5}
    \\
     MN10 & 84.2  & - & 91.1 & 87.4 & - & \textbf{93.8}
    \\
    \hline
  \end{tabular}
\end{table}

%

\begin{table}[!htb]
\centering
\caption{Retrieval and classification comparison in terms of Micro-averaged metrics under ShapeNetCore55.}  
    \begin{tabular}{c|c|c|c|c|c}  
     \hline
        & \multicolumn{5}{|c}{Micro} \\
     \hline
       Methods & P & R & F1 & mAP & NDCG  \\  
     \hline
       Kanezaki & 81.0 & 80.1 & \textbf{79.8} & 77.2 & 86.5 \\
       Zhou & 78.6 & 77.3 & 76.7 & 72.2 & 82.7 \\
       Tatsuma  & 76.5 & 80.3 & 77.2 & 74.9 & 82.8  \\
       Furuya  & \textbf{81.8} & 68.9 & 71.2 & 66.3 & 76.2  \\
       Thermos  & 74.3 & 67.7 & 69.2 & 62.2 & 73.2 \\
       Deng  & 41.8 & 71.7 & 47.9 & 54.0 & 65.4 \\
       Li  & 53.5 & 25.6 & 28.2 & 19.9 & 33.0 \\
       Mk  & 79.3 & 21.1 & 25.3 & 19.2 & 27.7 \\
       Su  & 77.0 & 77.0 & 76.4 & 73.5 & 81.5 \\
       Bai  & 70.6 & 69.5 & 68.9 & 64.0 & 76.5 \\
       Taco  & 70.1 & 71.1 & 69.9 & 67.6 & 75.6 \\
       Our  & 62.0 & \textbf{80.4} & 62.2 & \textbf{85.9} & \textbf{90.2} \\
        \hline
        \multicolumn{3}{c|}{SVSL\cite{Zhizhong2018seqijcai}} & \multicolumn{3}{|c}{85.5} \\
        \multicolumn{3}{c|}{VIPG\cite{Zhizhong2018VIPijcai}} & \multicolumn{3}{|c}{83.0} \\
       \multicolumn{3}{c|}{Our classification} & \multicolumn{3}{|c}{\textbf{86.9}} \\
     \hline
   \end{tabular}
   \label{table:t10}
\end{table}


\section{Conclusions }
Parts4Feature is proposed to learn 3D global features from part-level information in a semantic way. It successfully learns universal knowledge of generally semantic part detection from 3D segmentation benchmarks, and effectively transfers the knowledge to other shape analysis benchmarks by learning 3D global features from detected parts in multiple views. Parts4Feature makes it feasible to improve 3D global feature learning by leveraging discriminative information from another source. Moreover, our novel view aggregation with multi-attention can also benefit Parts4Feature to learn more discriminative features than widely used aggregation procedures. Our outperforming results show that Parts4Feature is superior to other state-of-the-art methods.

\section{Acknowledgments}
This work was supported by National Key R\&D Program of China (2018YFB0505400) and NSF under award number 1813583. We thank all anonymous reviewers for their constructive comments.

\bibliographystyle{named}
\bibliography{paper}

\end{document}